\documentclass[runningheads]{llncs}
\usepackage{graphicx}
\usepackage{comment}
\usepackage[dvipsnames]{xcolor}
\usepackage{amsmath,amssymb} 
\usepackage{booktabs}
\usepackage[final]{pdfpages}
\usepackage[sort,nocompress]{cite}
\usepackage{ulem}

\usepackage{hyperref}
\hypersetup{
	pdfborder={0 0 0 [0 0]}
}

\newcommand{\ra}[1]{\renewcommand{\arraystretch}{#1}}

\newcommand{\etal}{\textit{et al}. }
\newcommand{\eg}{\textit{e.g.} }

\begin{document}
\pagestyle{headings}
\mainmatter
\def\ECCVSubNumber{2530}  

\title{General 3D Room Layout from a Single View by Render-and-Compare} 


%

\author{Sinisa Stekovic\inst{1} \and
Shreyas Hampali\inst{1} \and
Mahdi Rad\inst{1} \and
Sayan Deb Sarkar\inst{1} \and
Friedrich Fraundorfer\inst{1} \and
Vincent Lepetit\inst{1,2}}


%
\authorrunning{S. Stekovic et al.}
%

\institute{Institute for Computer Graphics and Vision, Graz University of Technology, Graz, Austria \and
Universit\'e Paris-Est, \'Ecole des Ponts ParisTech, Paris, France \\
	{\tt\small \{sinisa.stekovic, hampali, rad, sayan.sarkar, fraundorfer, lepetit\}@icg.tugraz.at} \\ 
	Project page: \href{https://www.tugraz.at/index.php?id=40222}{ \color{blue} https://www.tugraz.at/index.php?id=40222}}

\maketitle

\begin{abstract}

We present a novel method  to  reconstruct the  3D layout  of a  room---walls, floors, ceilings---from a  single perspective view in challenging conditions, by contrast with previous single-view methods restricted to cuboid-shaped layouts.  This input view can consist of a color  image  only,  but  considering  a depth  map results in a more  accurate reconstruction.  Our approach is formalized as solving a constrained discrete optimization problem to find the set of 3D polygons that constitute the layout. In order to deal with occlusions between components of the layout, which is a problem ignored by previous works, we introduce  an analysis-by-synthesis method to iteratively refine the 3D layout estimate. As no dataset was available to evaluate our method quantitatively, we created one together with several appropriate metrics. Our dataset consists of 293 images from ScanNet, which we  annotated with precise 3D layouts. It offers three times more samples than the popular NYUv2 303 benchmark, and a much larger variety of layouts.
\end{abstract}


\renewcommand{\topfraction}{1}
\renewcommand{\dbltopfraction}{1}
\renewcommand{\bottomfraction}{1}
\renewcommand{\textfraction}{.0}
\renewcommand{\floatpagefraction}{1}
\renewcommand{\dblfloatpagefraction}{1}

\section{Introduction}

\begin{figure}
    \centering
    \begin{tabular}{ccccc}
        \includegraphics[height=.2\linewidth]{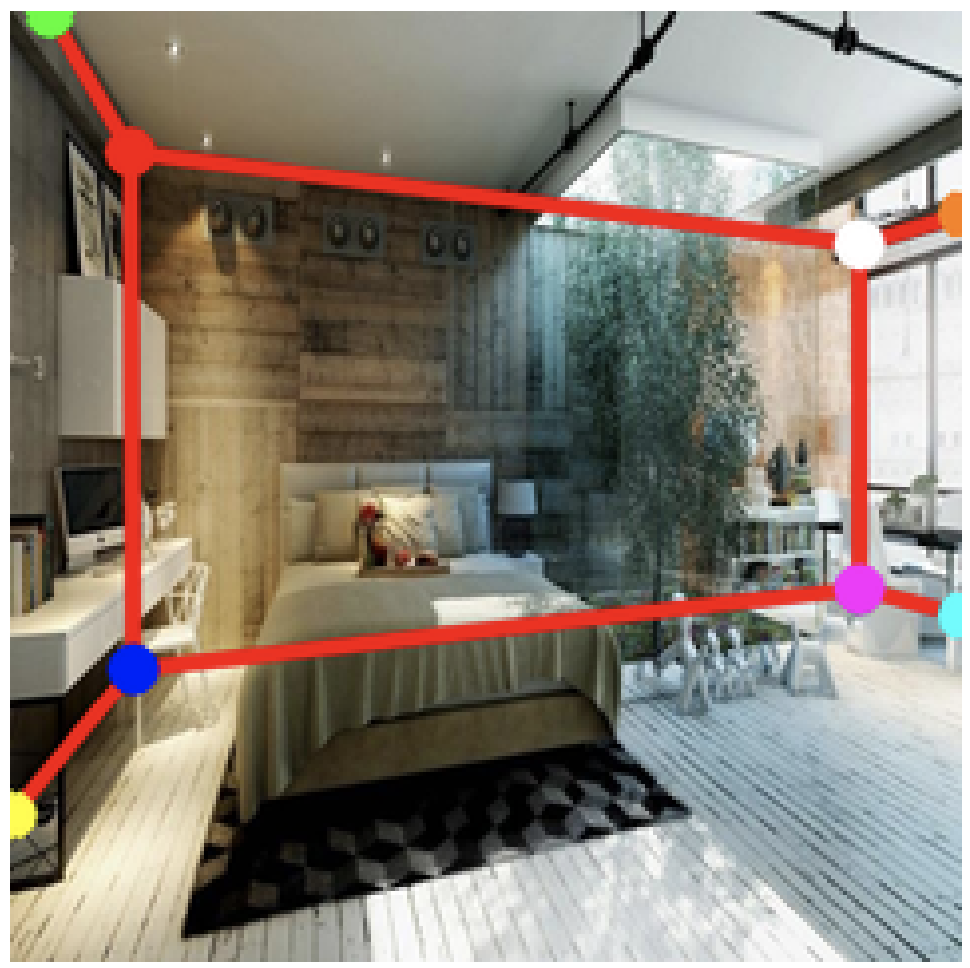} &
        \hspace{1cm} &
        \includegraphics[height=.2\linewidth]{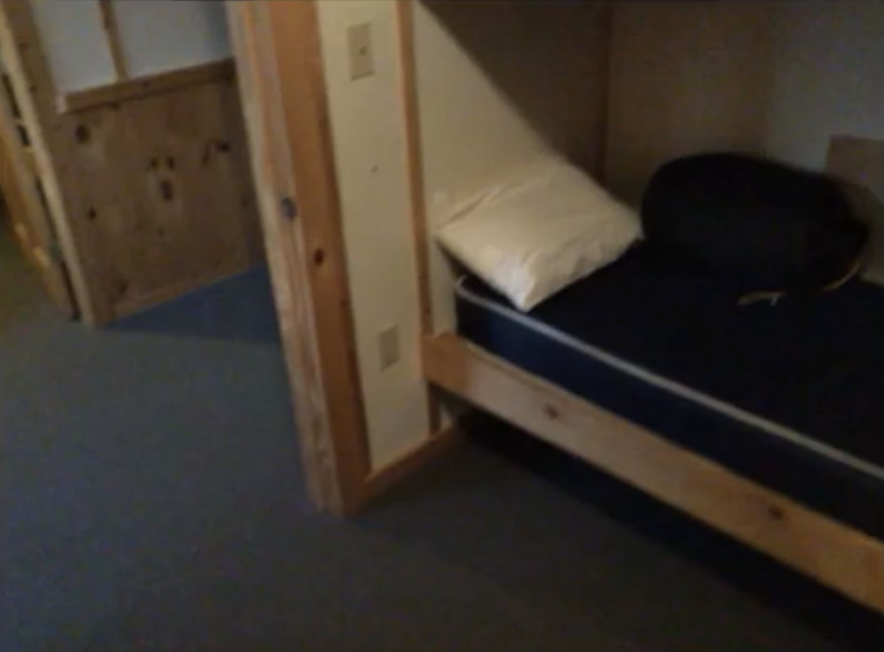} &
        \includegraphics[height=.2\linewidth]{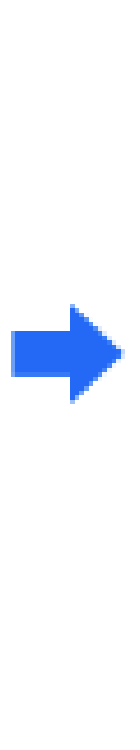} &
        \includegraphics[height=.2\linewidth]{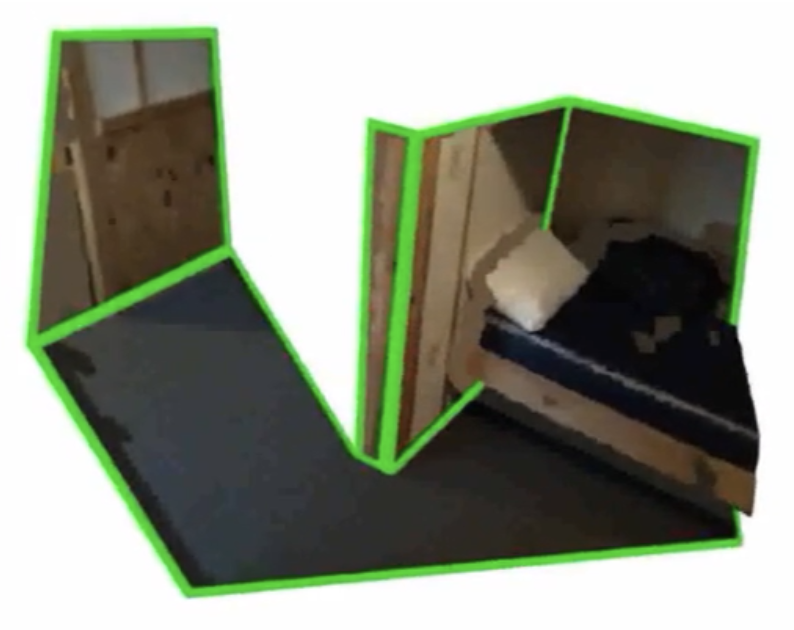} \\
        (a) A cuboid layout & & \multicolumn{3}{c}{(b) This paper} \\
    \end{tabular}
    \caption{(a) Most current methods for single view layout estimation make the assumption that the view contains a single room with a cuboid shape.  This makes the problem significantly simpler as the structure and number of corners remain fixed, but can only handle a fraction of indoor scenes. (b) By contrast, our method is able to estimate general 3D layouts from a single view, even in case of self-occlusions. Its input is either an RGBD image, or an RGB image from which a depth map is predicted. }
    \label{fig:teaser}
\end{figure}

The goal of layout estimation is to identify the layout components---floors, ceilings, walls--- and their 3D geometry from one or multiple views, despite the presence of clutter such as furniture, as illustrated in Fig.~\ref{fig:teaser}. This is a fundamental problem in scene understanding from images, with potential applications in many domains, including robotics and augmented reality. When enough images from different views are available, it is possible to recover complex 3D layouts by first building a dense point cloud~\cite{budroni2010automated,Liu2018floornet}. Single view scenarios are far more challenging even when depth information is available, since layout components may occlude each other, entirely or partially, and large chunks of the layout are then missing from the point cloud. Moreover, typical scenes contain furniture and the walls, the floors, and the ceilings might be obstructed. Important features such as corners or edges might be only partially observable or even not visible at all.

As shown in Fig.~\ref{fig:teaser}(a), many recent methods for single view scenarios avoid  these challenges by assuming that the room is a simple 3D cuboid~\cite{Schwing12,Mallya15,Lee2017roomnet,Holistic18,Zhang19,Hirzer2019smart,Chen2019holistic} or that the image contains at most 3 walls, a floor, and a ceiling~\cite{Zou2019complete}. This is a very strong assumption, which is not valid for many rooms or scenes, such as the ones depicted in Fig.~\ref{fig:teaser}(b) and Fig.~\ref{fig:our_results}. In addition, most of these methods only provide  the \textit{2D projection} of the layout~\cite{Schwing12,Lee2017roomnet,Zhang19,Hirzer2019smart}, which is not sufficient for many applications. Other methods rely  on panoramic images from viewpoints that do not create occlusions~\cite{Zou18,Sun2019horizonnet,Zou20193d}, which is not always feasible.

The very recent method by Howard-Jenkins~\etal~\cite{Howard2018outsidethebox} is probably the only method to be able to recover general layouts from a single perspective view. However, it does not provide quantitative evaluation for this task, but only for cuboid layouts in the case of single views. In fact, it often does not estimate well the extents of the layout components, and how they are connected together.

In this paper we introduce a formalization of the problem and an algorithm to solve it. Our algorithm takes as input a single view which can be an RGBD image, or even only a color image: When a depth map is not directly available, it is robust enough to rely on a predicted one from the color image~\cite{Liu2019planercnn,Ramamonjisoa2019sharpnet}. As shown on the right of Fig~\ref{fig:teaser}(b), its output is a 3D model that is "structured", in the sense that the layout components connected in the scene are also connected in the 3D model in the same way, similarly to what a human designer would do. Moreover, we introduce a novel dataset to quantitatively evaluate our method.

More exactly, we formalize the problem of recovering a 3D polygonal model of the layout as a constrained discrete optimization problem. This optimization selects the polygons constituting the layout from a large set of potential polygons. To generate these polygons, like \cite{Howard2018outsidethebox} and earlier work\cite{Nan2017polyfit} for point clouds, we rely on 3D planes rather than edges and/or corners to keep the approach simple in terms of perception and model creation. However, as  mentioned above, not all 3D planes required in the construction of the layout are visible in the image. Hence, we rely on an analysis-by-synthesis approach, sometimes referred to as 'render-and-compare'. Such approaches do not always require a realistic rendering, in terms of texture or lighting, as in \cite{Kundu20183d, Xu2019denserac} for example: We render a depth map for our current layout estimate, and compare it to the measured or predicted depth map. From the differences, we introduce some of the missing polygons to improve our layout estimate. We iterate this process until convergence. 

Our approach therefore combines machine learning and geometric reasoning. Applying "pure" machine learning to these types of problems is an appealing direction but it is challenging to rely only on machine learning to obtain structured 3D models as we do. Under the assumption that the room is box-shaped~\cite{Holistic18,Chen2019holistic}, this is possible because of the strong prior on feasible 2D layouts~\cite{Lee2017roomnet,Hirzer2019smart}. In the case of general layouts, this is difficult, as the variability of the layouts are almost infinite (see Fig.~\ref{fig:our_results} for examples). Moreover, only very limited annotated data is available for the general problem. Thus, we use machine learning only to extract image cues on the 3D layout from the perspective view, and geometric reasoning to adapt to general configurations based on these image cues.

To evaluate our method, we manually annotated 293 perspective views from the ScanNet test dataset~\cite{Dai2017} together with 5 novel 2D and 3D metrics, as there was no existing benchmark for the general problem. This is three times more images than NYUv2~303~\cite{Zhang2013estimating,Silberman2012}, a popular benchmark for evaluating cuboid layouts. Other single-view layout estimation benchmarks are Hedau~\etal~\cite{Hedau09} and LSUN~\cite{Zhang2015large}, that are cuboid datasets with only 2D annotations, and Structured3D~\cite{Structured3D}, a dataset containing a large number of synthetic scenes generated under the Manhattan world assumption. Our ScanNet-Layout dataset is therefore more general, and is publicly available. Table~\ref{table:scannet_dataset} summarizes the difference between benchmarks. We also compare our method to cuboid-specific methods on NYUv2~303, which contains only cuboids rooms, to show that our method performs comparably to these specialized methods while being more general.

\begin{table}[t]
    \centering
     \ra{1.3}
     \begin{tabular}{@{}rccrrcrcccrcr@{}} \toprule
    & Dataset & \multicolumn{2}{c}{Layout} & \multicolumn{2}{c}{Mode} & \multicolumn{2}{c}{Cam. Param.} & \multicolumn{2}{c}{Eval. Metrics} & \multicolumn{2}{c}{\#TestSamples} & \\
    & Hedau~\etal~\cite{Hedau09}  & Cuboid &&& RGB && Varying && 2D && 105 \\
    & LSUN~\cite{Zhang2015large}  & Cuboid &&& RGB && Varying && 2D && 1000 \\
    & NYUv2 303~\cite{Zhang2013estimating} & Cuboid &&& RGBD && Constant && 2D && 100 \\
    & ScanNet-Layout (ours) & General &&& RGBD && Constant && 3D && 293 \\
    \bottomrule
    \end{tabular}
    \vspace{0.5cm}
    \caption{Comparison between test sets of different datasets for single-view layout estimation on real images. ScanNet-Layout does not provide a training set.}
    \label{table:scannet_dataset}
\end{table}

\textbf{Main contributions}. First, we introduce a formalization of the general layout estimation from single views into  a constrained discrete optimization problem. Second, We propose an algorithm based on this formalization and are able to generate a simplistic 3D model for general layouts from single perspective views~(RGB or RGBD). Finally, we provide a novel benchmark with a dataset and new metrics for the evaluation of methods for general layout estimations from single perspective views.

\section{Related work}

We divide here previous works on layout estimation into two categories. Methods from the first category start by identifying features such as room corners or edges from the image. Like our own approach, methods from the second category rely on 3D planes and their intersections to build the layout. We discuss them below.

\subsection{Layout Generation from Image Features}
Some approaches to layout estimation, mostly for single-view scenarios, attempt to identify features in the image such as room corners or edges, before connecting them into a 2D layout or lifting them in 3D to generate a 3D room layout. Extracting such features, and lifting them in 3D are, however, very challenging. A common assumption is the Manhattan constraint that enforces orthogonality and parallelism between the layout components of the scene, often done by estimating vanishing points~\cite{Hedau09,Schwing12,Ramalingam13,Mallya15}, a process that can be very sensitive to noise. 

Another assumption used by most of the current methods is that only one box-shaped room is visible in the image~\cite{Schwing12, Zhang2013estimating,  Mallya15,Lee2017roomnet,Zhang19,Hirzer2019smart}. This is a strong prior that achieves good results, but only if the assumption is correct. For example, in~\cite{Mallya15}, 3D cuboid models are fitted to an edge map extracted from the image, starting from an initial hypothesis obtained from vanishing points. From this 3D cuboid assumption, RoomNet~\cite{Lee2017roomnet} defines a limited number of 11 possible 2D room layouts, and trains a CNN to detect the 2D corners of the box-shaped room. \cite{Zhang19} relies on segmentation to identify these corners more robustly. These last approaches~\cite{Lee2017roomnet,Zhang19} are limited to the recovery of a 2D cuboid layout.

Yet another approach is to directly predict the 3D layout from the image: \cite{Holistic18,Chen2019holistic} not only predict the layout but also the objects and humans present in the image, using them as additional constraints. Such constraints are very interesting, however, this approach also requires the `3D cuboid assumption', as it predicts the camera pose with respect to a 3D cuboid.

\cite{Zou18,Sun2019horizonnet,Zou20193d} relax the cuboid assumption and can recover more general layouts. However, in addition to the Manhattan assumption, this line of work requires panoramic images captured so that they do not exhibit occlusions from walls. This requirement can be difficult to fulfill or even impossible for some scenes. By contrast, our method does not require the  cuboid or the Manhattan assumptions, and handles occlusions in the input view to handle variety of general scenes.

\subsection{Layout Generation from 3D Planes}

An alternative to inferring room layouts from image features like room corners is to identify planes and infer the room layout from these plane structures. If complete point clouds are available, for example, from multiple RGB or RGBD images, identifying such planes is straightforward, and has been successfully applied for this task~\cite{sanchez2012planar, Cabral2014floorplan, ikehata2015structured, murali2017indoor}. Recently, successes in single RGB image based depth estimation~\cite{Godard2017monodepth,Ramamonjisoa2019sharpnet,Lee2019big} and 3D plane detection~\cite{Liu2019planercnn} opened up the possibility  of layout generation from single RGB images. For example, Zou~\etal~\cite{Zou2019complete} finds the layout planes in dominant room directions and then reasons on the depth map to estimate the extents of the layout components. However, even though this method does not assume strict cuboid layouts, it assumes the presence of only five layout components---floor, ceiling, left wall, front wall and right wall.

Like our method, the work of Howard-Jenkins~\etal~\cite{Howard2018outsidethebox} uses plane detected in images by a CNN to infer the non-cuboid 3D room layouts. The main contribution of their work is in the design of a network architecture to detect planar regions of the layout from a single image and to infer the 3D plane parameters from it. For this task, we use PlaneRCNN~\cite{Liu2019planercnn}, which has very similar functionalities. By intersecting these planes, they can be first delineated, and thanks to a clustering and voting scheme, with help of predicted bounding boxes for the planes, the parts of the planes relevant to the layout can be identified.

However, \cite{Howard2018outsidethebox} heavily relies on predicted proposal regions to estimate the extents of layout components. As their qualitative results show, it sometimes struggles to find the correct extents as the proposal regions can be very noisy, and the layout components can be disconnected. It also does not provide any quantitative evaluation for the general room layout estimation problem for single views and it is limited to cuboid rooms from the NYUv2~303 dataset~\cite{Zhang2013estimating}. 

In contrast, we formalize the problem as a discrete optimization problem, allowing us to reason about occlusions between layout components in the camera view, which often happens in practice, and retrieve structured 3D models.

\newcommand{\calC}{\mathcal{C}}
\newcommand{\calE}{\mathcal{E}}
\newcommand{\calS}{\mathcal{S}}
\newcommand{\calP}{\mathcal{P}}
\newcommand{\calR}{\mathcal{R}}
\newcommand{\calX}{\mathcal{X}}

\newcommand{\Cost}{K}
\newcommand{\cost}{k}
\newcommand{\ThD}{\text{3D}}
\newcommand{\TwD}{\text{2D}}
\newcommand{\bx}{{\bf x}}
\newcommand{\IoU}{\text{IoU}}
\newcommand{\PE}{\text{PE}}
\newcommand{\EE}{\text{EE}}
\newcommand{\gt}{\text{gt}}
\newcommand{\RMSE}{\text{RMSE}}
\newcommand{\uts}{\text{uts}}

\section{Approach}

\begin{figure}[t]
    \centering
    \setlength{\tabcolsep}{0.1pt}
    \begin{tabular}{ccccc}
        \multicolumn{5}{c}{\bf First iteration}\\
        \includegraphics[width=.19\linewidth]{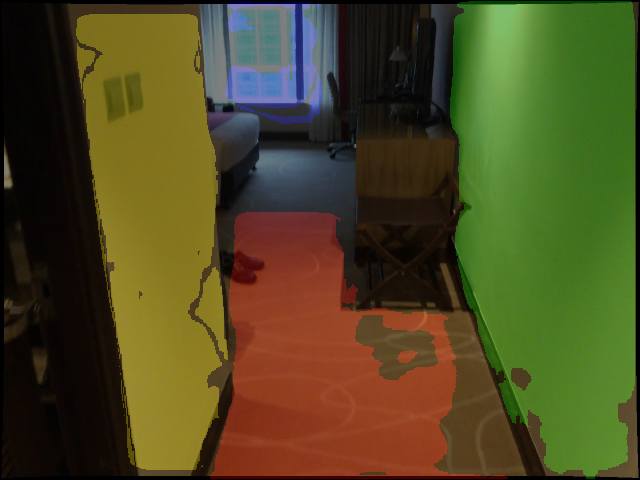}&
        \includegraphics[width=.19\linewidth]{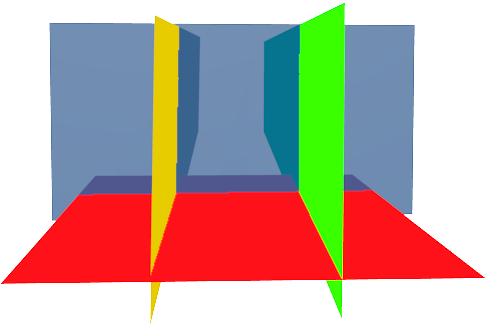}&
         \includegraphics[width=.19\linewidth]{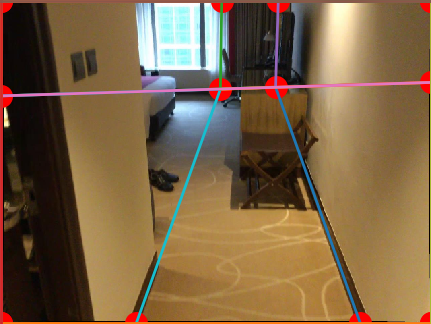}&
         \includegraphics[width=.19\linewidth]{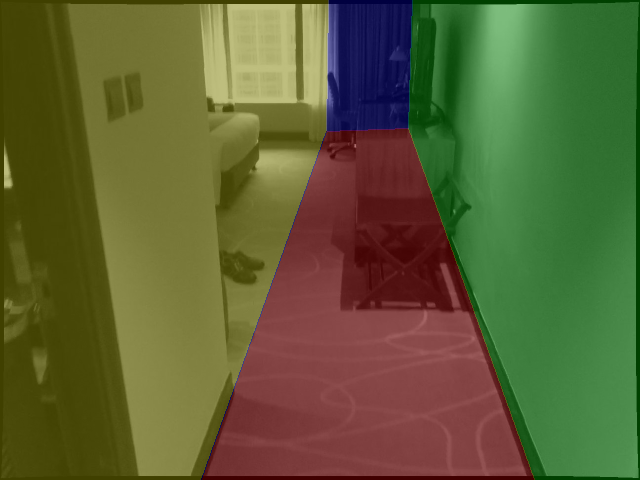}&
         \includegraphics[width=.19\linewidth]{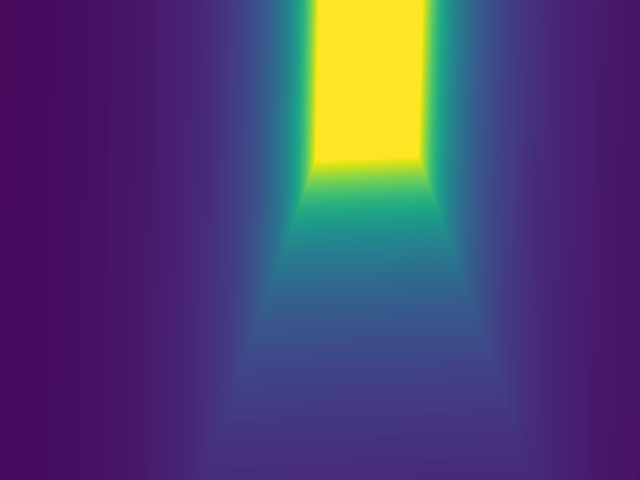}\\
         (a) planar regions & (b) 3D planes & (c) edges & (d) layout & (e) pred. depth\\[0.6em]
        \multicolumn{5}{c}{\bf Second iteration}\\
        \includegraphics[width=.19\linewidth]{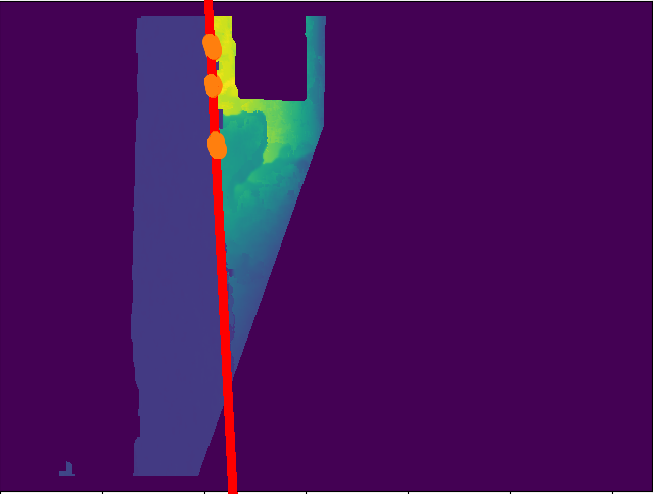}&
         \includegraphics[width=.19\linewidth]{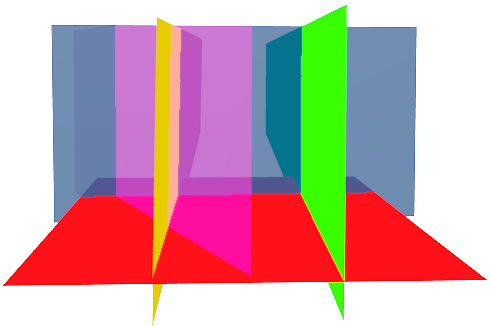}&
         \includegraphics[width=.19\linewidth]{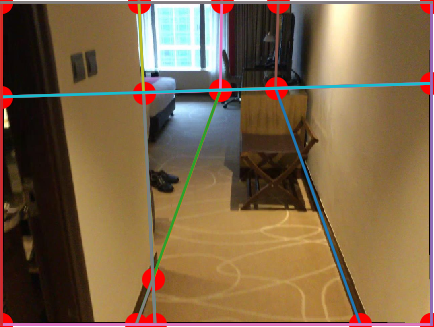}&
         \includegraphics[width=.19\linewidth]{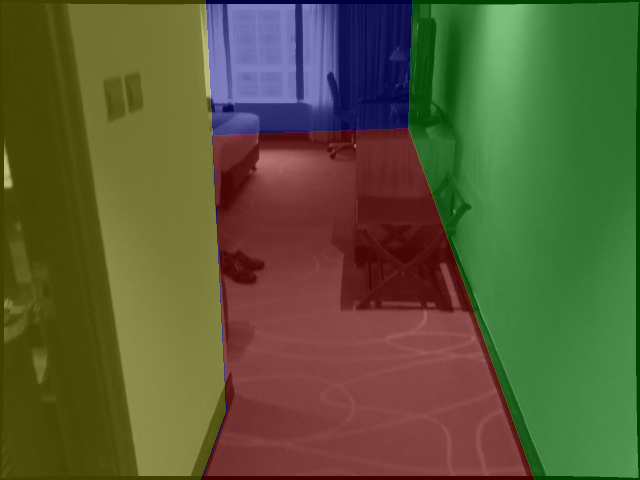}&
          \includegraphics[width=.19\linewidth]{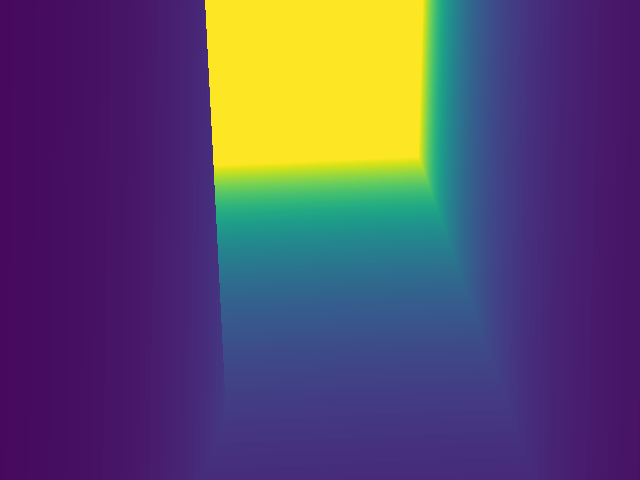}\\
        (f) fitted line & (g) new 3D planes & (h) new edges & (i) new layout & (j) pred. depth\\
    \end{tabular}
    \caption{Approach overview. We detect planar regions (a) for the layout components using PlaneRCNN and a semantic segmentation, and obtain equations of the corresponding 3D planes (b). The planes intersections give a set of candidate edges for the layout (c). From these edges, we find a first layout estimate in 2D~(d) and 3D~(e) as a set of polygons that minimizes the cost. From the depth discrepancy (f) for the layout estimate and the input view, we find missing planes (g), and extend the set of candidate edges (h). We iterate until we find a layout consistent with the color image (i) and the depth map (j).}
    \label{fig:overview}
\end{figure}

We describe our approach in this section. We formalize  the general layout estimation problem as a constrained discrete optimization problem~(Section~3.1), explain how we generate a first set of candidate polygons from plane intersections~(Section~3.2), detail our cost function involved in our formalization~(Section~3.3), and how we optimize it~(Section~3.4). When one or more walls are hidden in the image, this results in an imperfect layout, and we show how to augment the set of candidate  polygons to include these hidden walls, and iterate until we obtain the final layout~(Section~3.5). Finally, we describe how we can output a structured 3D model for the layout~(Section~3.6).

\subsection{Formalization}

We formalize the problem of estimating a 3D polygonal layout $\hat{\calR}$ for a given input image $I$ as solving the following constrained discrete optimization problem:
\begin{equation}
    \begin{array}{l}
    \hat{\calR} = \arg\min\limits_{\calX\subset \calR_0(I)} \Cost(\calX, I) \text{ such that } p(\calX) \text{ is a partition of }I \> ,
\end{array}
\label{eq:problem}
\end{equation}
where $\Cost(\calX, I)$ is a cost function defined below, $\calR_0(I)$ is a set of 3D polygons for image $I$, and $p(\calX)$ is the set of projections in the input view of the polygons in $\calX$. In words, we look for the subset of polygons in $\calR_0(I)$, whose projections partition the input image $I$, and that  minimizes $\Cost(\cdot)$.

There are two options when it comes to defining precisely $\Cost(\calX, I)$ and $\calR_0(I)$: Either $\calR_0(I)$ is defined as the set of all possible 3D polygons, and $\Cost(\calX,I)$ includes constraints to ensure that the polygons in $\calX$ reproject on image cues for the edges and corners of the rooms, or $\calR_0(I)$ contains only polygons with edges that correspond to edges of the room. As discussed in the introduction, extracting wall edges and corners from images is difficult in general, mostly because of lack of training data. We therefore chose the second option. We describe below first how we create the set $\calR_0(I)$ of candidate 3D polygons, which includes the polygons constituting the 3D layout, and then the cost function $\Cost(\calX,I)$.


\subsection{Set of Candidate 3D Polygons $\calR_0(I)$}
\label{sec:candidates}

As discussed in the introduction, we rely on the intersection of planes to identify good edge candidates to constitute the polygons of the layout. We then group these edges into polygons to create $\calR_0(I)$.

{\bf Set of 3D planes $\calP_0$.} First, we run on the RGB image a) PlaneRCNN~\cite{Liu2019planercnn} to detect planar regions and b) DeepLabv3+~\cite{Deeplabv3plus2018} to obtain a semantic segmentation. We keep only the planar regions that correspond to wall, ceiling, or floor segments (the supplementary material provides more details). We denote by $\calS(I)$ the set of such regions. An example is shown in Fig.~\ref{fig:overview}(a). PlaneRCNN provides the equations of the 3D planes it detects, or, if a depth map of the image is available, we fit a 3D plane to each detected region to obtain  more accurate parameters. The depth map can be measured or predicted from the input image $I$~\cite{Liu2019planercnn,Ramamonjisoa2019sharpnet}. 

As can be seen in Fig.~\ref{fig:overview}(a), the regions provided by PlaneRCNN typically do not extend to the full polygonal regions that constitute the layout.  To find these polygons, we rely on the intersections of the planes in $\calP_0$ as detailed below. In order to limit the extent of the polygons to the borders of the input image, we also include in $\calP_0$ the four 3D planes of the camera frustum, which pass through two neighbouring image corners and the camera center.

Some planes required to create some edges of the layout may not be in this first set $\calP_0$. This is the case for example for the plane of the hidden wall on the left of the scene in Fig.~\ref{fig:overview}. Through an analysis-by-synthesis approach, we can detect the absence of such planes, and add plausible planes to recover the missing edges and obtain the correct layout. This will be detailed in Section~\ref{sec:iterative}.

{\bf Set of 3D corners $\calC_0$.} By computing the intersections of each triplet of planes in $\calP_0$, we get a set $\calC_0$ of candidate 3D corners for the layout. To  build a structured layout, it is important to keep track of the planes that generated the corners and, thus, we define each corner $C_j\in\calC_0$ as a set of 3 planes:
\begin{equation}
C_j = \{P_j^1, P_j^2,P_j^3\} \> ,
\end{equation}
where $P_j^1\in\calP_0$, $P_j^2\in\calP_0$, $P_j^3\in\calP_0$, and $P_j^1\neq P_j^2$, $P_j^1\neq P_j^3$, and $P_j^2\neq P_j^3$. For numerical stability, we do not consider the cases where at least two planes are  almost parallel, or when the 3 planes almost intersect on a line. Furthermore, we discard the corners that reproject outside the image. We also discard those corners that have negative depth values. 

{\bf Set of 3D edges $\calE_0$.} We then obtain a set $\calE_0$ of candidate 3D edges by pairing the corners in $\calC_0$ that share exactly 2 planes:
\begin{equation}
E_k = \{C_{\sigma(k)}, C_{\sigma'(k)}\} \> ,
\end{equation}
where $\sigma(k)$ and $\sigma'(k)$ are 2 functions giving the indices of the corners that are the extremities of edge $E_k$. Fig.~\ref{fig:overview}(c) gives an example of set $\calE_0$. 

{\bf Set of 3D polygons $\calR_0(I)$.} We finally create the set $\calR_0(I)$ of candidate polygons as the set of all closed loops of edges in $\calE_0$ that lie on the same plane and do not intersect each other.


\subsection{Cost Function $\Cost(\calX, I)$}

Our cost function is split into a 3D and a 2D part:
\begin{equation}
    \Cost(\calX, I) = \Cost_\ThD(\calX, I) + \lambda \Cost_\TwD(\calX, I) \> .
\end{equation}
For all our experiments, we used $\lambda = 1$.

Cost function $K_\ThD(\cdot)$ measures the dissimilarity with the depth map $D(I)$ for the input view, and the depth map $D'(\calX)$ created from the polygons in $\calX$, as illustrated in Fig.~\ref{fig:overview}(e). It is based on the observation that the layout should always be located behind the objects of the scene:
\begin{equation}
\Cost_\ThD(\calX, I) = \frac{1}{|I|} \sum\limits_\bx \max(D(I)[\bx]-D'(X)[\bx], 0) \> ,\\
\end{equation}
where the sum is over all the image locations $\bx$ and $|I|$ denotes the total number of image locations. Since the projections of the polygons in $\calX$ are constrained to form a partition of $I$, $\Cost_\ThD(\cdot)$ can be rewritten as
\begin{equation}
\Cost_\ThD(\calX, I) = \frac{1}{|I|} \sum\limits_{R\in \calX} \sum\limits_{\bx\in p(R)} \max(D(I)[\bx]-D'(\calX)[\bx], 0) 
= \frac{1}{|I|} \sum\limits_{R\in \calX} \cost_\ThD(R, I) \> ,\\
\end{equation}
where $p(R)$ is the projection  of polygon $R$ in the image. $\Cost_\ThD(\cdot)$ is computed as a sum of terms, each term depending on a single polygon in $\calX$. These terms are precomputed for each polygon  in $\calR_0(I)$, to speed up the computation of  $\Cost_\ThD(\cdot)$.

Cost function $\Cost_\TwD(\cdot)$ measures the dissimilarity between the polygons in $\calX$ and the image segmentation into planar regions $\calS(I)$:
\begin{equation}
\begin{array}{rcl}
\Cost_\TwD(\calX, I) &=& \sum\limits_{R\in \calX} \> \Big(\big(1 - \IoU(p(R), S(I,R))\big) + \IoU(p(R), \calS(I) \setminus S(I,R)) \Big)\\
&=& \sum\limits_{R\in \calX} \> \cost_\TwD(R, I) \> ,\\
\end{array}
\end{equation}
where $\IoU$ is the Intersection over Union score, $S(I,R)$ is the planar region detected by Plane-RCNN and corresponding to the plane of polygon $R$. Like $\Cost_\ThD(\cdot)$, $\Cost_\TwD(\cdot)$ can be computed as a sum of terms that can be precomputed before optimization. Computing cost function $\Cost(\cdot)$ is therefore very fast.


\subsection{Optimization}

To find the solution to our constrained discrete optimization problem introduced in Eq.~\eqref{eq:problem}, we simply consider all the possible subsets $\calX$ in $\calR_0(I)$ that pass the partition constraint, and keep the one that minimizes $\Cost(\calX, I)$.

The number $N$ of polygons in $\calR_0(I)$ varies with the scene, but is typically of a few tens. For example, we obtain $12$ candidate polygons in total for the example of Fig.~\ref{fig:overview}. The number of non-empty subsets to evaluate is theoretically $2^{N}-1$, which  is slightly higher than $2000$ for the same example. However, most of these subsets can be trivially discarded: Associating polygons with corresponding planes and considering that only one polygon per plane is possible significantly reduces the number of possibilities, to $36$ in this example. The number can be further reduced by removing the polygons that do not  have a plausible shape to  be part of a room layout. Such shapes can be easily recognized by considering the distance between the non-touching edges of the polygon. Finally, this  reduces the number to merely $20$ plausible subsets of polygons in the case of the example. Precomputing the $\cost_\ThD$ and $\cost_\TwD$ terms takes about 1s, and the optimization itself takes about 400ms---most of this time is spent to guarantee the partition constraint in our implementation, which we believe can be significantly improved. The supplementary material details the computation time more.


\begin{figure}[t]
    \centering
    \setlength{\tabcolsep}{1pt}
    \begin{tabular}{cccc}
         \includegraphics[width=0.24\linewidth]{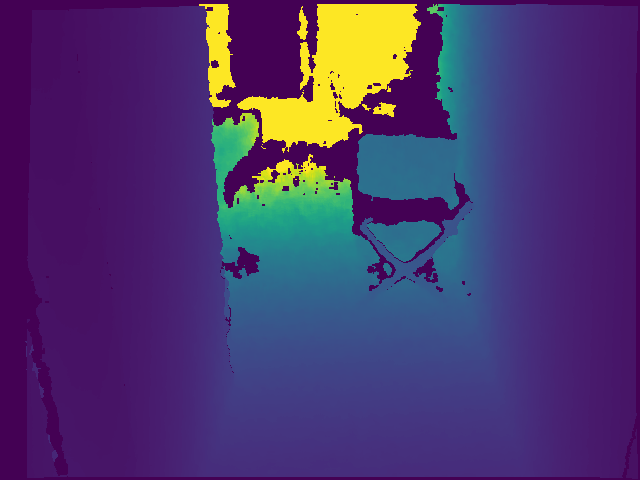} &
         \includegraphics[width=0.24\linewidth]{figures/method/overview3/5_2_layout_depth_iter0.png} &
         \includegraphics[width=0.24\linewidth]{figures/method/overview3/6_0_discr_line_fitting.png} &
         \includegraphics[width=0.24\linewidth]{figures/method/overview3/6_2_layout_depth_iter1.png} \\
         (a) & (b) & (c) & (d) \\
    \end{tabular}
    \caption{Layout Refinement. We identify planes which are occluded by other layout planes but necessary for the computation of the layout. First, we compare the depth map for the input view (a) to the rendered layout depth (b). (c) If the discrepancy is large, we fit a line (shown in red) through the points with the largest discrepancy change (orange). By computing the plane passing through the line and the camera center, we obtain a layout~(d) consistent with the depth map for the input view. }
    \label{fig:discr_lay}
\end{figure}

\subsection{Iterative Layout Refinement}
\label{sec:iterative}

As mentioned above in Section~\ref{sec:candidates}, we often encounter cases where some of the planes required to create the layout are not in $\calP_0$ because they are hidden by another layout plane. Fortunately, we can detect such mistakes, and fix them by adding a plane to $\calP_0$ before running the layout creation described above again.


To detect missing planes, we render the depth map $D'(\hat{\calR})$ for the current layout estimate $\hat{\calR}$ and measure the discrepancy with the depth map $D(I)$ for the image as illustrated in Fig.~\ref{fig:discr_lay}. As the depth maps $D(I)$ acquired by RGBD cameras typically contain holes around edges, we use the depth completion method by \cite{Ku2018defense} before measuring the discrepancy. If discrepancy is large, \textit{i.e.} there are many pixel locations where the rendered map has smaller values than the original depth map, this indicates a mistake in the layout estimate that can be fixed by adding a plane. This is because the layout cannot be in front of objects. 

There is a range of planes that can improve the layout estimate. We chose the conservative option that does not introduce parts not visible in the input image. For a polygon $R$ in $\hat{\calR}$ with a large difference between $D'(\hat{\calR})$ and $D(I)$, we first identify the image locations with the largest discrepancy changes, and fit a line to these points using RANSAC, as shown in Fig.~\ref{fig:overview}(f). We then add the plane $P$ that passes through this line and the camera center to $\calP_0$ to obtain a new set of planes $\calP_1$. This is illustrated in Fig.~\ref{fig:overview}(g): the intersection between $P$ and $R$ will create the edge missing from  the layout, which is visible in Fig.~\ref{fig:overview}(h). From $\calP_1$, we obtain successively the new sets $\calC_1$~(corners), $\calE_1$~(edges), and  $\calR_1$~(polygons), and solve again the problem of Eq.~\eqref{eq:problem} after replacing $\calR_0$ by $\calR_1$. We repeat this process until we do not improve the differences between $D'(\hat{\calR})$ and $D(I)$, for the image locations segmented as layout components.

For about $5\%$ of the test samples, the floor plane is not visible because of occlusions by some furniture. When none of the detected planes belongs to the floor class, we create an additional plane by assuming that the camera is $1.5m$ above the floor. For the plane normal, we take the average of the outer products between the normals of the walls and the $[0,0,1]^\top$ vector. 

\subsection{Structured Output}
Once the solution $\hat{\calR}$ to Eq.~\eqref{eq:problem} is found, it is straightforward to create a structured 3D model for the layout. Each 3D polygon in $\hat{\calR}$ is defined as a set of coplanar 3D edges, each edge is defined as a pair of corners, and each corner is defined from 3 planes. We therefore know which corners and edges the polygons share and 
how they are connected to each other. For example, the 3D layout of Fig.~\ref{fig:teaser} is made of 14 corners, 18 edges, and 5 polygons.
\section{Evaluation}

We evaluate our approach in this section. First, we present our new benchmark for evaluating 3D room layouts from single perspective views, and our proposed metrics. Second, we evaluate our approach on our benchmark and include both quantitative and qualitative results on general room layouts. For reference, we show that our approach performs similarly to methods assuming cuboid layouts on the NYUv2 303 benchmark, which only includes cuboid layouts, without making such strong assumptions. More qualitative results, detailed computation times, and implementation details are given in the supplementary material.

We have considered additionally evaluating our approach on the LSUN~\cite{Zhang2015large} and the Hedau~\cite{Hedau09} room layout benchmarks. However, because these datasets do not provide the camera intrinsic parameters, these datasets were unpractical for the evaluation of our approach. We have also considered evaluating our approach on 3D layout annotations of the NYUv2 dataset~\cite{Silberman2012} from \cite{Guo2013support}. However, as the annotations are not publicly available anymore, we were not able to produce any quantitative results for this dataset. Furthermore, the improved annotations from \cite{Zou2019complete} are not publicly available anymore, and the authors were unfortunately not able to provide these annotations in time for this submission.

\newcommand{\figres}{0.19\linewidth}
    
\begin{figure}[t]
    \centering
    \begin{tabular}{ccccc}
    \includegraphics[width=\figres]{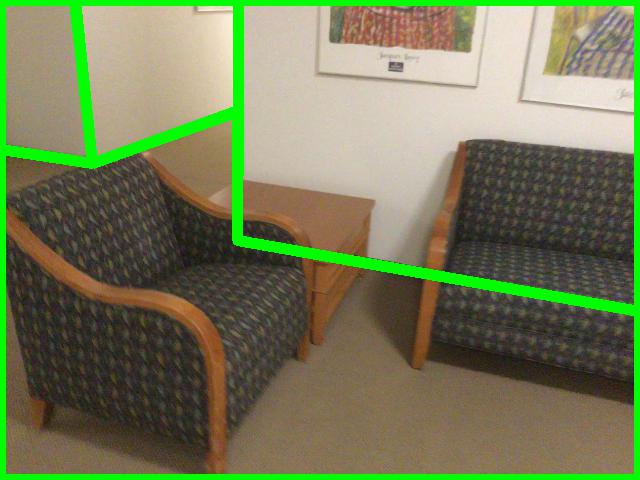} &
    \includegraphics[width=\figres]{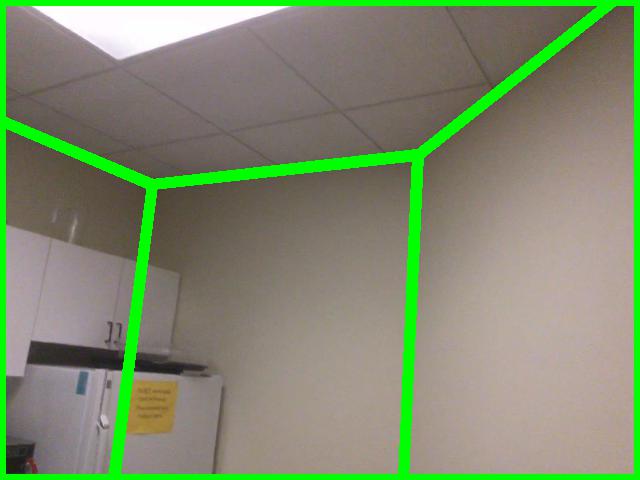} &
    \includegraphics[width=\figres]{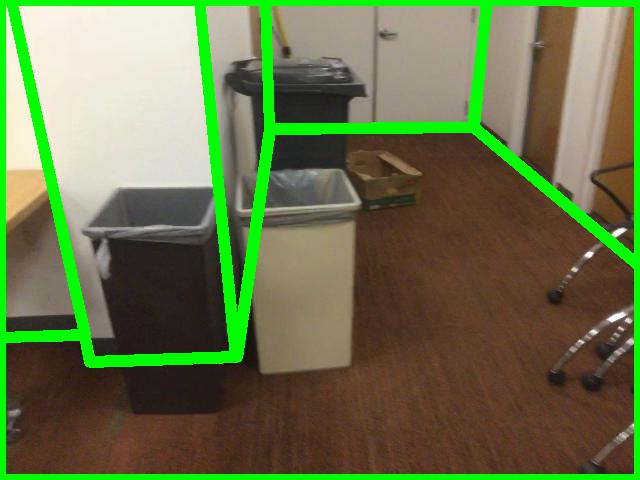} &
    \includegraphics[width=\figres]{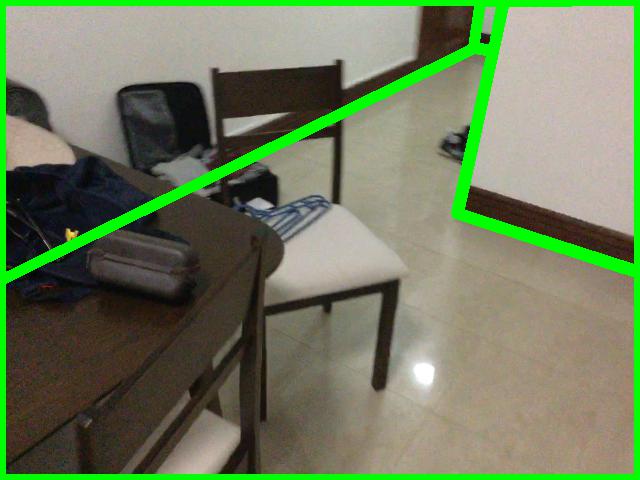} &
    \includegraphics[width=\figres]{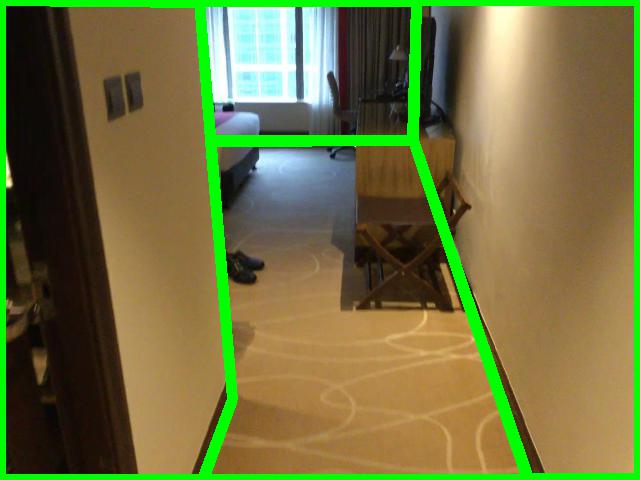} \\
    \includegraphics[width=\figres]{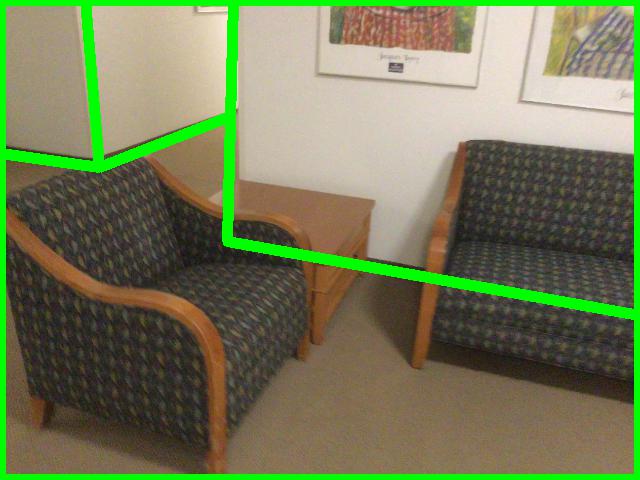} &
    \includegraphics[width=\figres]{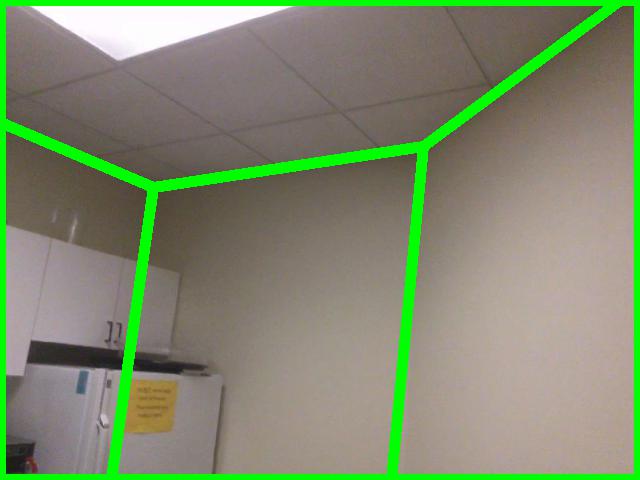} &
    \includegraphics[width=\figres]{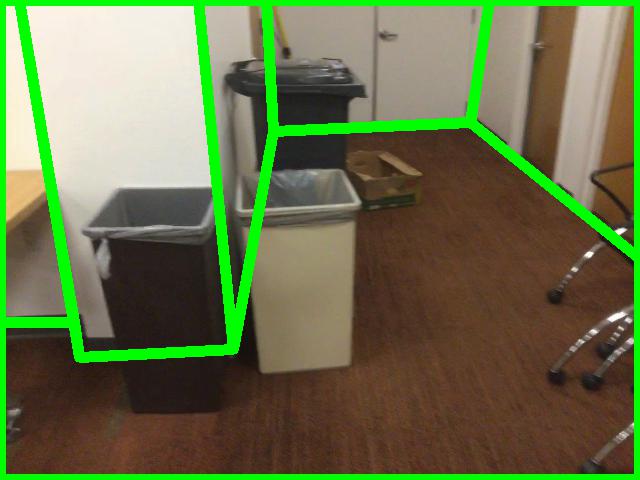} &
    \includegraphics[width=\figres]{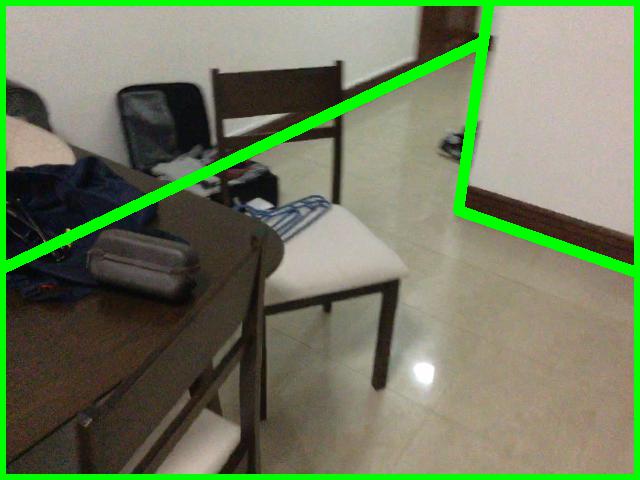} &
    \includegraphics[width=\figres]{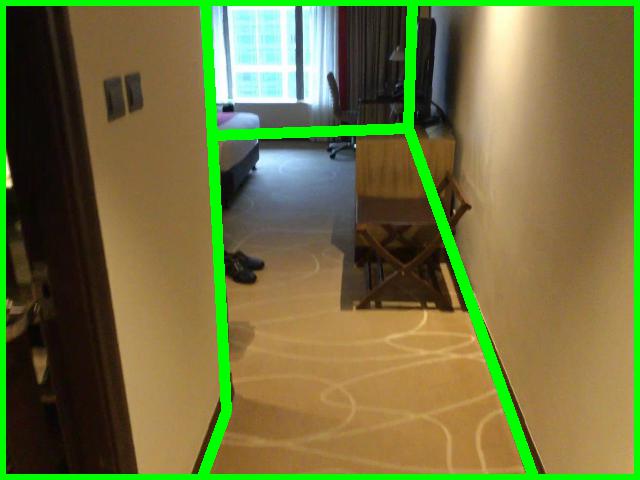} \\
    \includegraphics[width=\figres]{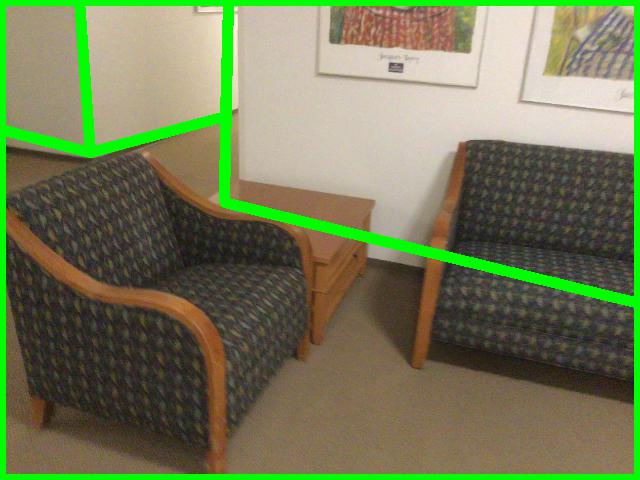} &
    \includegraphics[width=\figres]{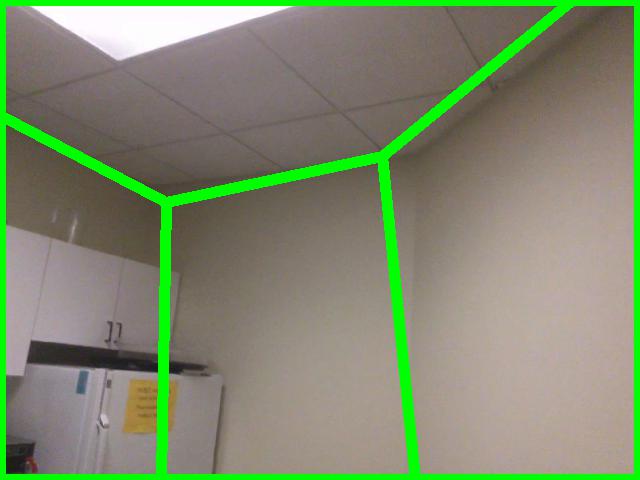} &
    \includegraphics[width=\figres]{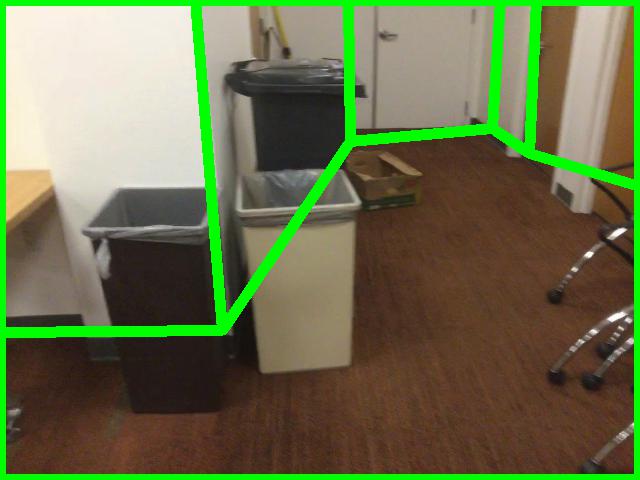} &
    \includegraphics[width=\figres]{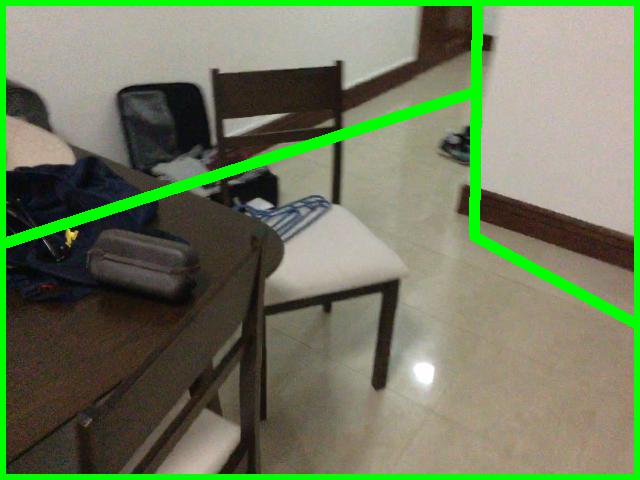} &
    \includegraphics[width=\figres]{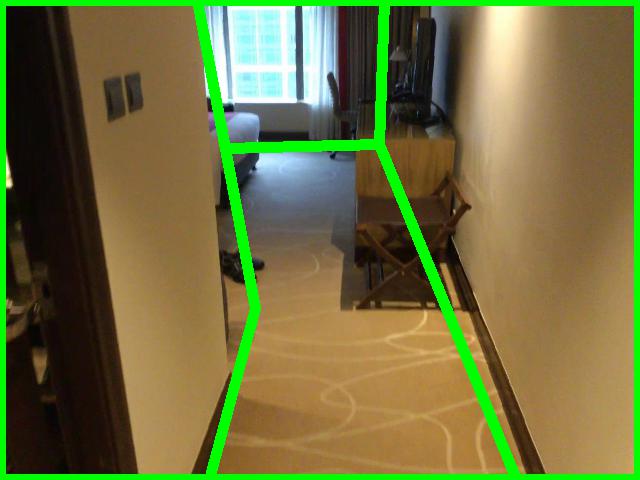} \\
    \includegraphics[width=\figres]{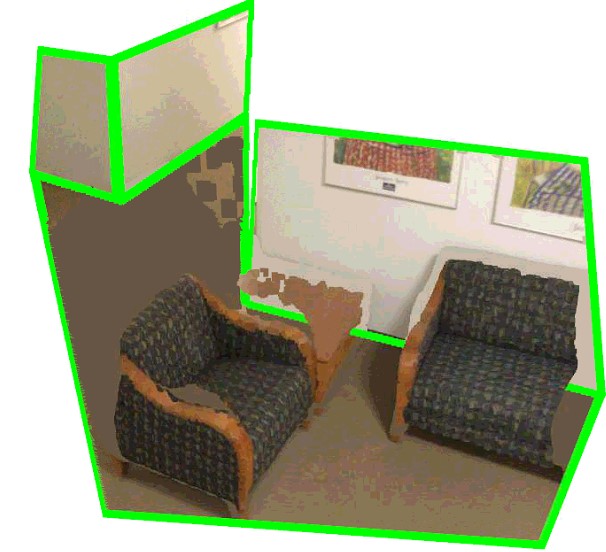} &
    \includegraphics[width=\figres]{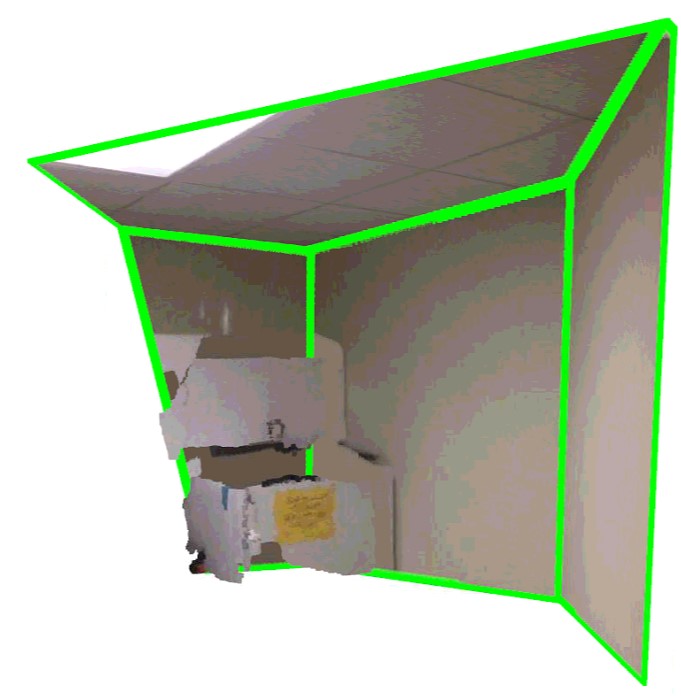} &
    \includegraphics[width=\figres]{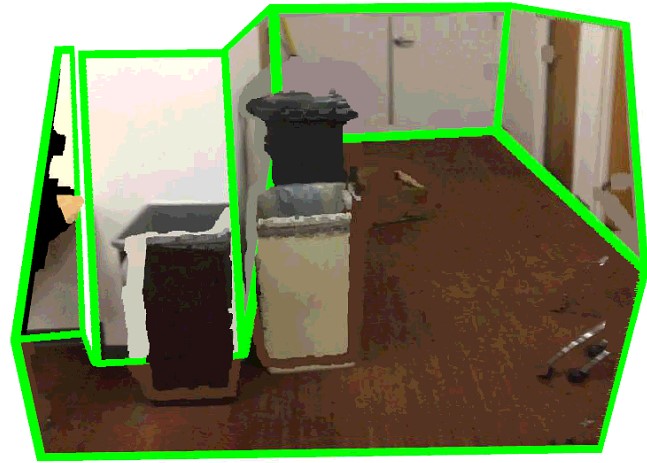} &
    \includegraphics[width=\figres]{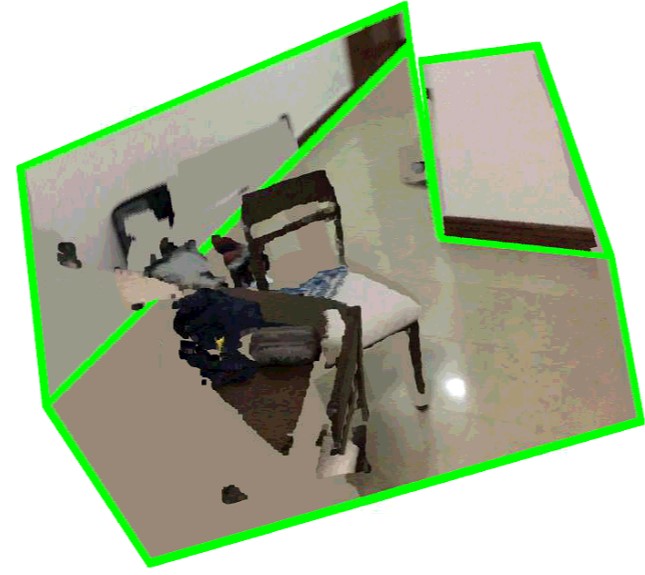} &
    \includegraphics[width=\figres]{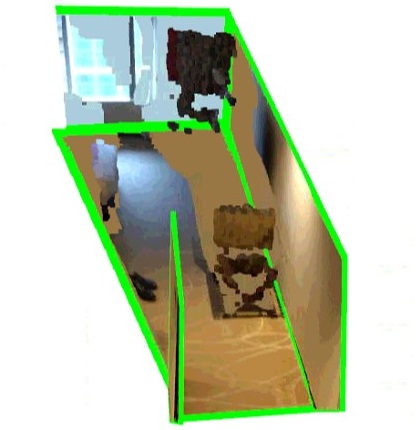}
    \\
    \end{tabular}
    \caption{Results of our method on the ScanNet-Layout. First row: Manual annotations; Second row: Predictions using an RGBD input; Third row: Predictions using an RGB input. Fourth row: 3D models created using the RGBD mode of our approach. Furniture is shown only to demonstrate consistency of our predictions with the geometry of the scene. Our approach performs well in both RGBD and RGB modes. Our approach in RGB mode fails to detect one component in the third example, due to noisy predictions from PlaneRCNN. The rest of the examples show that, when depth information is not available, predictions from CNN can still be utilized in many different scenarios. More qualitative results, including a video, can be found in the supplementary material.}
    \label{fig:our_results}
\end{figure}

\subsection{ScanNet-Layout Benchmark}

\textbf{Dataset creation.} For our ScanNet-Layout dataset, we manually labelled $293$ views sampled from the $100$ ScanNet test scenes~\cite{Dai2017}, for testing purposes only. As shown in Fig.~\ref{fig:our_results} and in the supplementary material, these views span different layout settings, are equally distributed to represent both cuboid and general room layouts, challenging views that are neglected in previous room layout datasets, and in some cases we include similar viewpoints to evaluate effects of noise~(\eg motion blur). The ScanNet-Layout dataset is available on our project page.

To manually annotate the 3D layouts, we first drew the layout components as 2D polygons. For each polygon, we then annotated the image region where it is directly visible without any occlusions from objects or other planes. From these regions, we could compute the 3D plane equations for the polygons. Since we could not recover the plane parameters for completely occluded layout components, we only provide 2D polygons without 3D annotations for them.

\textbf{Evaluation metrics.} To quantitatively evaluate the fidelity of the recovered layout structures and their 2D and 3D accuracy,  we introduce 2D and 3D metrics. For the 2D metrics, we first establish one-to-one correspondences $\calC$ between the $N$ predicted polygons $\hat{\calR}$ and the $M$ ground truth polygons $\calR_\gt$. Starting with the largest ground truth polygon, we iteratively find the matching predicted polygon with highest intersection over union. At each iteration, we remove the ground truth polygon and its match from further consideration. The metrics are:
\begin{itemize}
\item Intersection over Union~(IoU): $\frac{2}{M + N} \sum_{(R_\gt, R) \in \calC}  \IoU(R_\gt, R)$, where $\IoU$ is the Intersection-over-Union measure between the projections of the 2 polygons. This metric is very demanding on the global structure and 2D accuracy;
\item Pixel Error~(PE): $\frac{1}{|I|} \sum_{\bx \in I}  \PE(\bx)$, with $\PE(\bx) = 0$  if the ground truth polygon and the predicted polygon projected at image location $\bx$ were matched together, and 1  otherwise. This metric also evaluates the global structure;
\item Edge Error~(EE): This is the symmetrical Chamfer distance~\cite{Olson97b} between the polygons in $\hat{\calR}$ and $\calR_\gt$, and evaluates the accuracy of the layout in 2D;
\item Root Mean Square Error~(RMSE) between the predicted layout depth $D(\calR)$ and the ground truth layout depth $D(\calR_\gt)$, excluding the pixels that lie on completely occluded layout components, as we could not recover 3D data for these components. This metric evaluates the accuracy of the 3D layout.
\item $\RMSE_\uts$ that computes the RMSE after scaling the predicted layout depth to the range of ground truth layout depth by factor $s = \text{median}(D(\calR_\gt))\>/$  $\text{median}(D(\calR))$. This metric is used when the depth map is predicted from the image, as the scale of depth prediction methods is not reliable. 
\end{itemize}

We note that the PE and EE metrics are extensions of existing metrics in cuboid layout benchmarks. As the PE metric is forgiving when missing out small components, we introduce the IoU metric that drastically penalizes such errors.

\begin{table}[t]
    \centering
     \ra{1.3}
     \begin{tabular}{@{}rccrrcrrrcrcc@{}} \toprule
    & Mode & \multicolumn{2}{c}{$\IoU\uparrow$ (\%)} & \multicolumn{2}{c}{$\PE\downarrow$ (\%)} & \multicolumn{2}{c}{$\EE\downarrow$} & \multicolumn{2}{c}{$\RMSE\downarrow$} & \multicolumn{2}{c}{$\RMSE_\uts\downarrow$} & \\
    \hline
     Hirzer~\cite{Hirzer2019smart}  & RGB & $48.6 \pm 22.2$  & & $24.1 \pm 15.1$ & & $29.6 \pm 19.4$ & & - & & - &\\
     \textit{Ours}  & RGB & ${\it 63.5 \pm 25.2}$  & & ${\it 16.3 \pm 14.7}$& & ${\it 22.3 \pm 14.9}$ & & ${\it 0.5 \pm 0.5}$ & & ${\it 0.4 \pm 0.5}$ & \\
    \textit{Ours}  & RGBD & ${\bf 75.9 \pm 23.4}$  & & ${\bf 9.9 \pm 12.9}$& & ${\bf 11.9 \pm 13.2}$ & & ${\bf 0.2 \pm 0.4}$ & & ${\bf 0.2 \pm 0.3}$ \\
    \bottomrule
    \end{tabular}
    \vspace{0.5cm}
    \caption{Quantitative results on our ScanNet-Layout benchmark. ($\uparrow$: higher values are better, $\downarrow$: lower values are better) The numbers for Hirzer~\etal~\cite{Hirzer2019smart} demonstrate that the approaches assuming cuboid layouts under-perform on our ScanNet-Layout benchmark. Our approach in RGB performs much better as it is not restricted by these assumptions. Our approach in RGBD mode shows even more improvement.}
    \label{table:quant_scannet}
\end{table}

\subsection{Evaluation on ScanNet-Layout}

We evaluate our method on ScanNet-Layout under two different experimental settings: When depth  information is directly measured by a depth camera, and when only the color image is available as input. In this case, we use PlaneRCNN~\cite{Liu2019planercnn} to estimate both the planes parameters and the depth map.

\begin{figure}[t]
    \centering
    \begin{tabular}{ccccc}
    \includegraphics[width=0.23\linewidth]{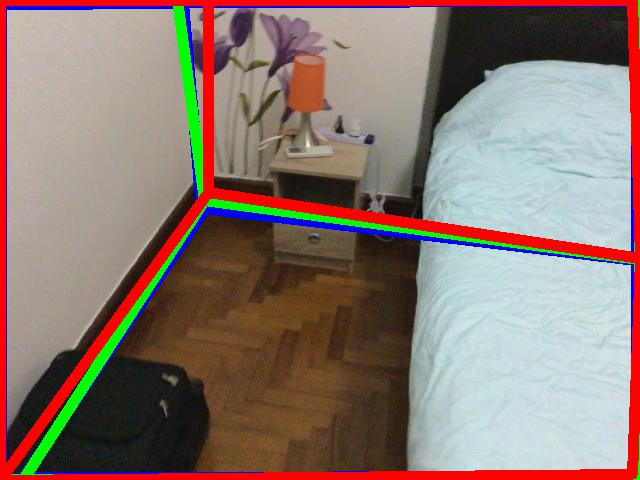} &
    \includegraphics[width=0.23\linewidth]{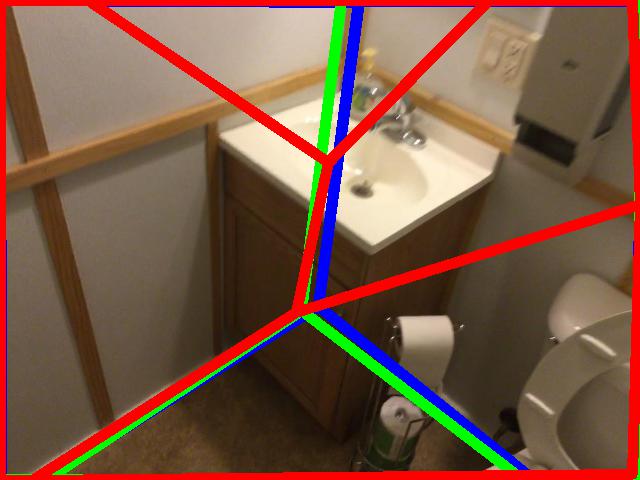} &
    \includegraphics[width=0.23\linewidth]{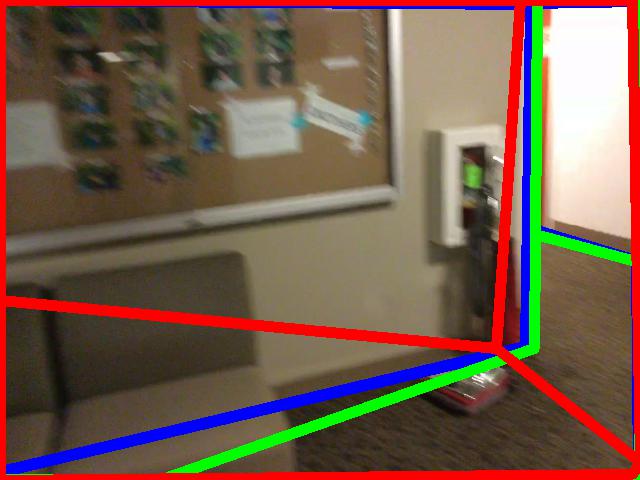} &
    \includegraphics[width=0.23\linewidth]{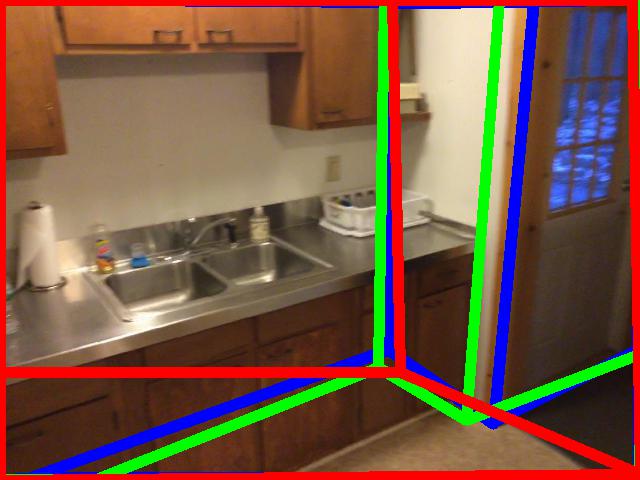} \\
    (a) & (b) & (c) & (d)\\
    \end{tabular}
    \caption{Visual comparison to Hirzer~\etal~\cite{Hirzer2019smart}, which assumes only cuboid layouts. The layouts estimated by the Hirzer method are shown in red, the layouts recovered by our approach using RGB information only are shown in green, and ground truth is shown in blue. (a) Both approaches perform similarly. (b) The Hirzer method makes a mistake even for this cuboid layout. (c) and (d) The Hirzer method fails as the cuboid assumption does not hold. Our approach performs well in all of the examples.}
    \label{fig:comp_hirzer}
\end{figure}

Table~\ref{table:quant_scannet} reports the quantitative results. 
The authors of \cite{Hirzer2019smart}, one of the state-of-the-art methods for cuboid layout estimation, kindly provided us with the results of their method. As this method is specifically designed for cuboid layouts, it fails on more general cases, but also on many cuboid examples for viewpoints not well represented in the training sets of layout benchmarks~\cite{Hedau09, Zhang2013estimating, Zhang2015large}~(Fig.~\ref{fig:comp_hirzer}(b)).

The good performance of our method for all metrics shows that the layouts are recovered accurately in 2D and 3D. When measured depth is not used, performance decreases due to noisy estimates of the plane parameters, but  in many cases, the predictions are still accurate. This can be best observed in qualitative comparisons with RGBD and RGB views in Fig.~\ref{fig:our_results}. In many cases, RGB information is enough to estimate 3D layouts and are comparable to results with RGBD information. However, the third example clearly demonstrates that small errors in planes parameters can lead to visible errors.

\begin{table}[t]
    \centering
    \ra{1}
     \begin{tabular}{@{}rcrcrccr@{}} \toprule
    &$\quad$& Mode &$\quad$& $\PE\downarrow$ &$\>$& $\text{Median } \PE\downarrow$ \\
    \hline
    Zhang~\etal~\cite{Zhang2013estimating} && RGBD && $8.04$ && - \\
    \textit{Ours} && RGBD && $8.9$ && $4.6$ \\
    \hline
    Schwing~\etal~\cite{Schwing12} && RGB && $13.66$ && - \\
    Zhang~\etal~\cite{Zhang2013estimating} && RGB && $13.94$ && - \\ 
    RoomNet~\cite{Lee2017roomnet} (from \cite{Hirzer2019smart})&& RGB && $12.31$ && - \\
    Hirzer \etal \cite{Hirzer2019smart} && RGB && $8.49$ && - \\
    Howard-Jenkins~\etal~\cite{Howard2018outsidethebox} && RGB && $12.19$ && - \\
    \textit{Ours} && RGB && $13.0$ && $10.1$ \\

    \bottomrule
    \end{tabular}
    \vspace{0.5cm}
    \caption{Quantitative results on NYUv2 303, a standard benchmark for cuboid room layout estimation. Our method performs similarly to the other methods designed for cuboid rooms without using this assumption. While  Hirzer~\etal~\cite{Hirzer2019smart} performs best on this benchmark, it fails on ScanNet-Layout, even for some of the cuboid rooms~(Fig.~\ref{fig:comp_hirzer}).}
    \label{table:quant_nyu}
\end{table}

\begin{figure}[t]
    \centering
    \newcommand{\figwidthnyu}{0.2}
    \begin{tabular}{ccc}
        \includegraphics[width=\figwidthnyu\linewidth]{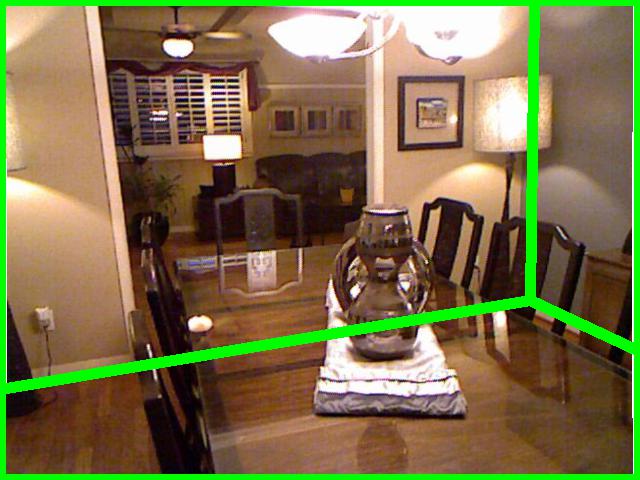} &
        \includegraphics[width=\figwidthnyu\linewidth]{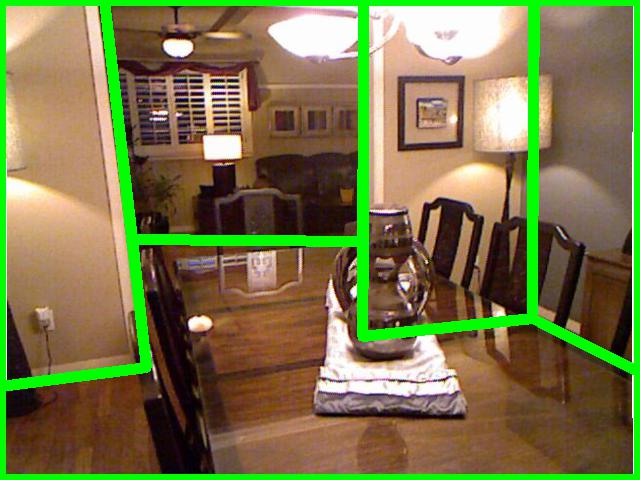}&
        \includegraphics[width=\figwidthnyu\linewidth]{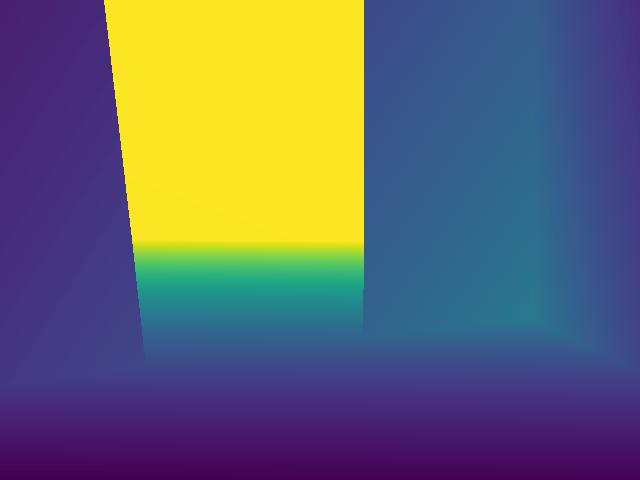}\\

        (a) & (b) & (c)\\
    \end{tabular}
    \centering
    \caption{Qualitative result on NYUv2 303. (a) Layout obtained by enforcing the cuboid assumption, (b) the original layout retrieved by our method, which corresponds better to the scenes. (c) Depth map computed from our estimated layout.}
    
    \label{fig:qual_nyu}
\end{figure}

\begin{figure}[t]
    \centering
    \begin{tabular}{cccc}
    \includegraphics[width=0.23\linewidth]{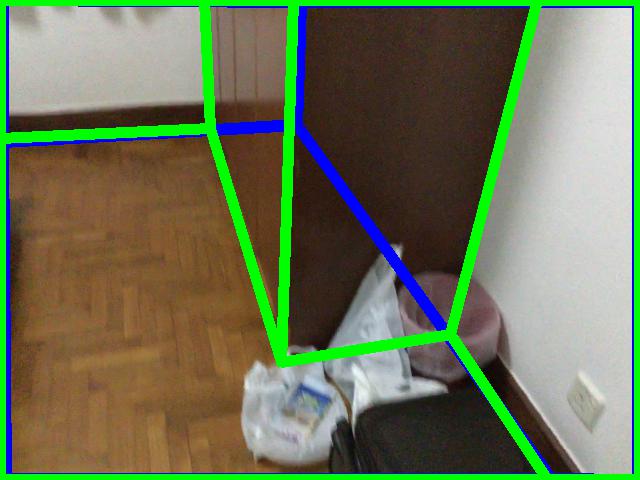} &
   \includegraphics[width=0.23\linewidth]{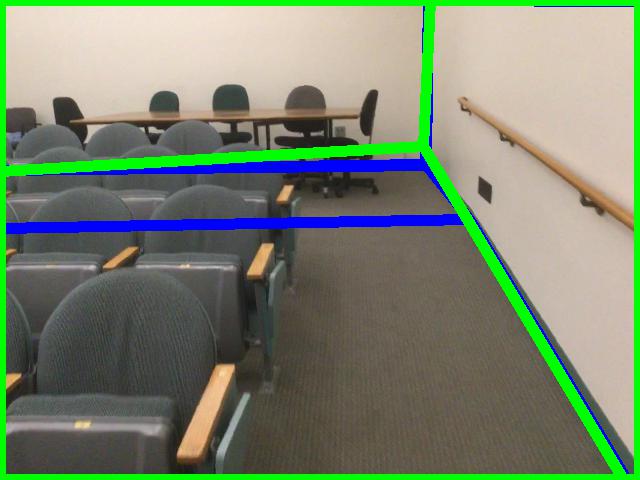} &
   \includegraphics[width=0.23\linewidth]{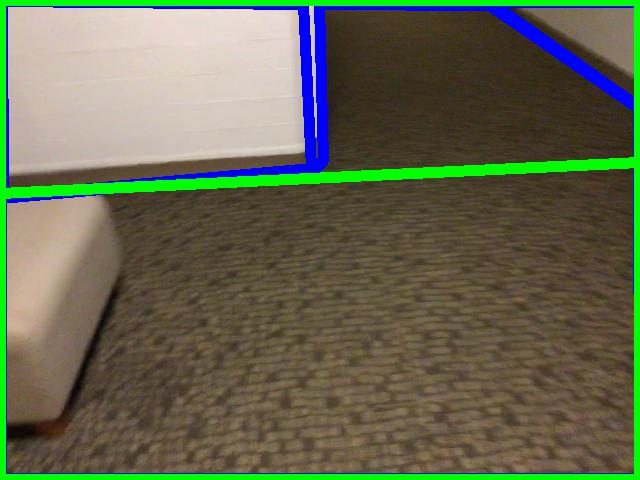} &
   \includegraphics[width=0.23\linewidth]{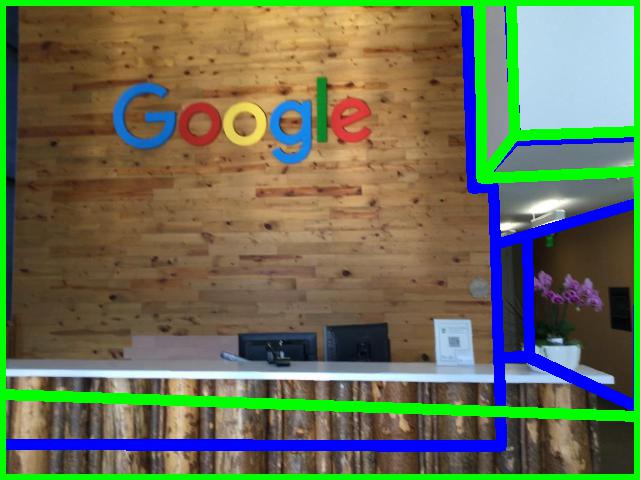}
    \\
    (a) & (b) & (c) & (d)  \\
    \end{tabular}
    \caption{Failure cases on ScanNet-Layout, our estimations in green, ground truth in blue. (a): Some furniture were segmented as walls. (b): PlaneRCNN did not detect the second floor plane. Even human observers may fail to see this plane. (c): Large areas in the measured depth map were missing along edges. Filling these areas with \cite{Ku2018defense} is not always sufficient to detect discrepancy. (d): As the floor is not visible in the image, it is unclear whether the manual annotation or the estimated floor polygon is correct.}
    \label{fig:fails}
\end{figure}

\subsection{Evaluation on NYUv2 303}

For reference, we evaluate our approach on the NYUv2 303 dataset~\cite{Zhang2013estimating,Silberman2012}. It is designed to evaluate methods predicting 2D room layouts under the cuboid assumption. We show that our method also performs well on it without exploiting this assumption.  
This dataset only provides annotations for the room corners under the cuboid assumption. Since the output of our method is more general, we transform it into the format expected by the  NYUv2 303 benchmark. For each of the possible cuboid layout components---1 floor, 1 ceiling, 3 walls---we find the planes for which its normal vector best fits the layout component:  When fewer than 3 walls are visible, the annotations of walls in the dataset are ambiguous and we apply the Hungarian algorithm~\cite{Kuhn55thehungarian} to find good correspondences.

Table~\ref{table:quant_nyu} gives the quantitative results. When depth is available, our method is slightly worse than the Zhang~\etal~\cite{Zhang2013estimating} method, designed for cuboid rooms. When using only color images, our method performs similarly to the other approaches, specialized for cuboid rooms, even if the recent Hirzer~\etal~\cite{Hirzer2019smart} method performs best on this dataset. Here, we used the depth maps predicted by \cite{Ramamonjisoa2019sharpnet} to estimate the plane parameters. Fig.~\ref{fig:qual_nyu} shows that  some layouts we retrieved fit the scene better than the manually annotated layout. To conclude, our method performs closely but is more general than the cuboid-based methods.

\subsection{Failure Cases}
Fig.~\ref{fig:fails} shows the frequent causes of failures. Most of the failures are due to noisy outputs from PlaneRCNN and DeepLabv3+ that lead to both false-positive and missing layout planes. Our render-and-compare approach is not robust enough to large noise in depth and this should be addressed in future work.

\section{Conclusion}

We presented a formalization of the general room layout estimation into a constrained discrete optimization problem, and an algorithm to solve this problem. The occasional errors made by our method come from the detection of the planar regions, the semantic segmentation, and the predicted depth maps, pointing to the fact that future progress in these fields will improve our layout estimates.

\textbf{\newline Acknowledgment.}This work was supported by the Christian Doppler Laboratory for Semantic 3D Computer Vision, funded in part by Qualcomm Inc.

\clearpage
%
%
\bibliographystyle{splncs04}
\bibliography{string,egbib}

\end{document}